\pdfoutput=1
\documentclass[11pt]{article}

\usepackage[final]{acl}

\usepackage{times}
\usepackage{latexsym}
\usepackage[T1]{fontenc}
\usepackage[utf8]{inputenc}
\usepackage{microtype}
\usepackage{inconsolata}
\usepackage{tabularx}
\usepackage{multirow}
\usepackage{graphicx}

\title{Detecting Reference Errors in Scientific Literature\\
with Large Language Models
}

\author{Tianmai M. Zhang \quad Neil F. Abernethy \\
University of Washington\\
\texttt{\{tianmai, neila\}@uw.edu}}

\begin{document}
\maketitle
\begin{abstract}
Reference errors, such as citation and quotation errors, are common in scientific papers. Such errors can result in the propagation of inaccurate information, but are difficult and time-consuming to detect, posing a significant challenge to scientific publishing. To support automatic detection of reference errors, this work evaluated the ability of large language models in OpenAI’s GPT family to detect quotation errors. Specifically, we prepared an expert-annotated, general-domain dataset of statement-reference pairs from journal articles. Large language models were evaluated in different settings with varying amounts of reference information provided by retrieval augmentation. Our results showed that large language models are able to detect erroneous citations with limited context and without fine-tuning. This study contributes to the growing literature that seeks to utilize artificial intelligence to assist in the writing, reviewing, and publishing of scientific papers. Potential avenues for further improvements in this task are also discussed.
\end{abstract}

\section{Background and Introduction}

Researchers cite literature as references and supporting evidence when reporting their work in papers. The reliability of referencing is usually taken for granted. However, previous citation verification studies in multiple scientific domains have revealed that reference errors of varying degrees are common in scientific papers, with prevalence rates ranging from 11\% to 41\%, depending on the domain, journal, and methodology (\citealp{Goldberg93, Lee99, Fenton00, Gosling04, Lukic04, Todd10, Jergas15, Mogull17, Armstrong18, Smith20, Pavlovic21, Cobb24}). Reference errors could result in the propagation of inaccurate information \citep{Smith20}, undermining the credibility of scientific research and sometimes leading to serious consequences \citep{Pavlovic21}. For example, hundreds of uncritical citations of a 1980 letter published in the New England Journal of Medicine may have contributed to the opioid crisis in the United States \citep{Leung17}.

Previous studies defined two major types of reference errors: citation errors and quotation errors. Citation errors usually refer to typographical errors in referencing, such as incorrect reference information (e.g., incorrect authors, title, journal, or year) or the erroneous arrangement of references \citep{Smith20}. Citation errors have become less common in the era of digitization and citation managers, although these same factors may enable the propagation of pre-existing citation errors. In contrast, a quotation error specifically refers to the situation where a reference fails to support the statement for which it is cited \citep{Smith20}. Notably, these two definitions are not mutually exclusive, as an incorrect assignment of reference indices can manifest as a quotation error in an individual statement-reference pair. Recent studies on referencing errors sometimes use different terms for quotation errors, such as “content errors” \citep{Mogull17}, “inaccurate citations” \citep{Pavlovic21}, or “miscitations” \citep{Cobb24}.

Quotation errors are difficult and time-consuming to detect, and they often require domain expertise when comparing a statement to relevant information in the reference article \citep{Smith20}. Previous studies on reference errors typically utilized domain experts to manually examine samples of scientific papers. The difficulty of detecting quotation errors poses a significant challenge to scientific publishing, as it requires additional efforts by editors and reviewers in peer review. Given the exploding number of papers being published each year, this task is becoming increasingly demanding.

Recent advances in natural language processing (NLP) have demonstrated astounding capabilities of large language models (LLMs) to perform various types of text-based tasks (\citealp{Ouyang22, Touvron23, OpenAI24, Gemini24}), providing a strong baseline for application in the real world. Researchers have also started exploring ways NLP can assist with paper writing and peer review \citep{Kuznetsov24}. However, none of these studies examined reference error detection. To fill this gap and encourage future attempts to automate reference error detection, this study performed a general-domain evaluation of the capability of LLMs to detect quotation errors in scientific papers.

\section{Task and Related Work}

The quotation error detection task of this work is defined as follows: given a statement $s$ and a reference article $r$ that the statement cites, a model should predict a label $f(s,r)\in$ \{Fully substantiated, Partially substantiated, Unsubstantiated\} to indicate whether the statement-reference pair contains a quotation error. The names and definitions of the labels follow previous citation verification studies \citep{Smith20, Cobb24}. Complete definitions of the labels are listed in Table~\ref{tab:label_definitions}.

\begin{table}
  \centering
  \begin{tabularx}{\columnwidth}{X}
    \hline
    \textbf{Fully substantiated}: The reference article fully substantiates the relevant part of the statement.\\
    \hline
    \textbf{Partially substantiated}: According to the reference article, there is a minor error in the statement, but the error does not invalidate the purpose of the statement.\\
    \hline
    \textbf{Unsubstantiated}: The reference part does not substantiate any part of the statement. This could be because the statement is contradictory to, unrelated to, or simply missing from the reference article.\\
    \hline
  \end{tabularx}
  \caption{Label definitions.}
  \label{tab:label_definitions}
\end{table}

There have been some studies in the computer science domain that both utilized citations and employed a similar task or methodology (to ours) but were for different scenarios. Motivated by fact-checking claims related to COVID-19, \citet{Wadden20} proposed a scientific claim verification task in which each atomic statement/claim is verified against a corpus of paper abstracts to determine whether it is supported or refuted by the literature. Several teams of researchers, including Wadden et al. themselves, subsequently developed models for this shared task using the same dataset (\citealp{Pradeep21, Li21, Zhang21, Wadden22}). Notably, the scientific claim verification task labels a claim that is neither supported nor refuted by the corpus as “Not Enough Information”. Furthermore, the scope of the scientific claim verification task was limited to biomedicine. In terms of LLM-generated citations, Wu et al. examined the quality of web-based, instead of peer-reviewed article based, citations generated by LLMs in response to medical questions \citep{Wu24}.

\section{Dataset and Experiments}

The evaluation dataset is available on GitHub\footnote{\url{https://github.com/tianmai-zhang/ReferenceErrorDetection}}. Distributions of labels, domains, and reference availability in the dataset are summarized in Table~\ref{tab:dataset_characteristics}. Some examples of quotation errors in the dataset as well as data collection and quality control details are provided in Appendix~\ref{sec:quotation_error_example} and~\ref{sec:data_collection_details}, respectively.

\begin{table}
  \centering
  \begin{tabularx}{\columnwidth}{ll}
    \hline
    \textbf{Label} & \textbf{n (\%)}\\
    \hline
    Unsubstantiated & 112 (44.8)\\
    Partially substantiated & 14 (5.6)\\
    Fully substantiated & 124 (49.6)\\
    \hline
    \textbf{Domain} & \textbf{n (\%)}\\
    \hline
    Biology or Medicine & 85 (34.0)\\
    Chemistry or Material Science & 57 (22.8)\\
    Physics & 26 (10.4)\\
    Social Science & 26 (10.4)\\
    Earth or Environmental Science & 24 (9.6)\\
    Engineering & 17 (6.8)\\
    Computer Science & 15 (6.0)\\
    \hline
    \textbf{Reference Availability} & \textbf{n (\%)}\\
    \hline
    Has abstract & 242 (96.8)\\
    Has PDF & 244 (97.6)\\
    Has abstract or PDF & 250 (100)\\
    \hline
  \end{tabularx}
  \caption{Dataset characteristics.}
  \label{tab:dataset_characteristics}
\end{table}

LLMs were given task instructions and information about both the citing article and the reference. The prompt template (Appendix~\ref{sec:prompt_template}) was finalized before the start of the experiment. To investigate the impact of information completeness, LLMs were tested in 3 settings with different amounts of information from the reference: (1) With only the title of the reference provided; (2) With both the title and abstract provided; (3) With the title, abstract, and excerpts provided.

Local retrieval of excerpts from the main body of a reference followed a 3-step retrieval-augmented generation (RAG) \citep{Gao24} pipeline. First, the full text of a reference was extracted from its PDF file by GROBID\footnote{\url{https://github.com/kermitt2/grobid}}. The extracted full text was then split into 256-token chunks with 20-token overlaps using LlamaIndex\footnote{\url{https://github.com/jerryjliu/llama_index}}. In the experiment, the embeddings of the chunks were compared to that of the input statement using LlamaIndex, and the top 3 chunks with the highest match to the statement were retrieved and included in the prompt to the LLMs.

Additionally, we included OpenAI's Assistant API in the experiment since it provides a proprietary RAG workflow for LLMs to utilize information in PDF attachments. The assistant received the same prompt format containing the title of the reference, plus instructions to see the attached PDF file for the full text of the reference.

Three LLMs in OpenAI's GPT family were evaluated in the experiment: \texttt{gpt-3.5-turbo-0125}, \texttt{gpt-4-0125-preview}, and \texttt{gpt-4o-2024-05-13}. LLMs were prompted to respond with a JSON object containing a predicted label and an explanation for their selection. All LLM experiments were conducted using OpenAI's Python API with temperature set to 0. Model performance was measured by label accuracy. Since \citeposs{Wadden20} scientific claim verification task also applies to single statement-reference pairs, relevant models were also tested on our dataset for comparison.

\section{Results}

\begin{table*}
  \centering
  \begin{tabular}{llcccc}
    \hline
    \textbf{Model} & \textbf{Given Information} & \multicolumn{3}{c}{\textbf{Accuracy by Class}} & \textbf{Accuracy}\\
    & & \textbf{Un} & \textbf{Partially} & \textbf{Fully} & \textbf{Overall}\\
    \hline
    GPT-3.5 Turbo & Title & 64.3 & 14.3 & 73.4 & 66.0\\
    & Title + Abstract & 84.8 & 57.1 & 30.6 & 56.4\\
    & Title + Abstract + Excerpts & 79.5 & 57.1 & 30.6 & 54.0\\
    & Title + PDF (Assistant) & 79.5 & 14.3 & 63.7 & 68.0\\
    \hline
    GPT-4 Turbo & Title & 89.3 & 14.3 & 36.3 & 58.8\\
    & Title + Abstract & 89.3 & 35.7 & 39.5 & 61.6\\
    & Title + Abstract + Excerpts & 83.9 & 21.4 & 62.9 & 70.0\\
    & Title + PDF (Assistant) & 84.8 & 35.7 & 55.6 & 67.6\\
    \hline
    GPT-4o & Title & 90.2 & 14.3 & 29.8 & 56.0\\
    & Title + Abstract & 92.0 & 28.6 & 13.7 & 49.6\\
    & Title + Abstract + Excerpts & 86.6 & 50.0 & 34.7 & 58.8\\
    & Title + PDF (Assistant) & 83.9 & 21.4 & 58.9 & 68.0\\
    \hline
  \end{tabular}
  \caption{LLM performance on quotation error detection.}
  \label{tab:results_all}
\end{table*}

Table~\ref{tab:results_all} provides by-class and overall performance of the models in different settings. GPT-4 Turbo and GPT-4o performed much better than GPT-3.5 Turbo at detecting Unsubstantiated cases, especially when information about the reference article is limited. Provision of more information or the Assistant pipeline did not necessarily improve LLM performance, especially for GPT-3.5 Turbo which performed much better on Fully substantiated cases in the title-only setting than with more information. The models for \citeposs{Wadden20} scientific claim verification task predicted all statement-reference pairs in our dataset as “Not Enough Information” when using abstracts as references, suggesting that these models may not be directly applicable to our quotation error detection task. Therefore, they are excluded from Table~\ref{tab:results_all}.

Error analysis revealed that LLMs with low accuracy on Fully substantiated pairs predicted most of them as Partially substantiated. This was frequently the case when a statement contained multiple sub-claims, and LLMs were “confused” by the ambiguity in the span of the part to be verified. For instance, the statement in error example 1 of Appendix~\ref{sec:llm_error_examples} cites the reference article to support its data from South China only. GPT-4 Turbo assumed that the reference should substantiate the entire statement and predicted this case as Partially substantiated because the reference does not support other locations mentioned (Greenland and Svalbard).

\begin{figure}[t]
  \includegraphics[width=\columnwidth]{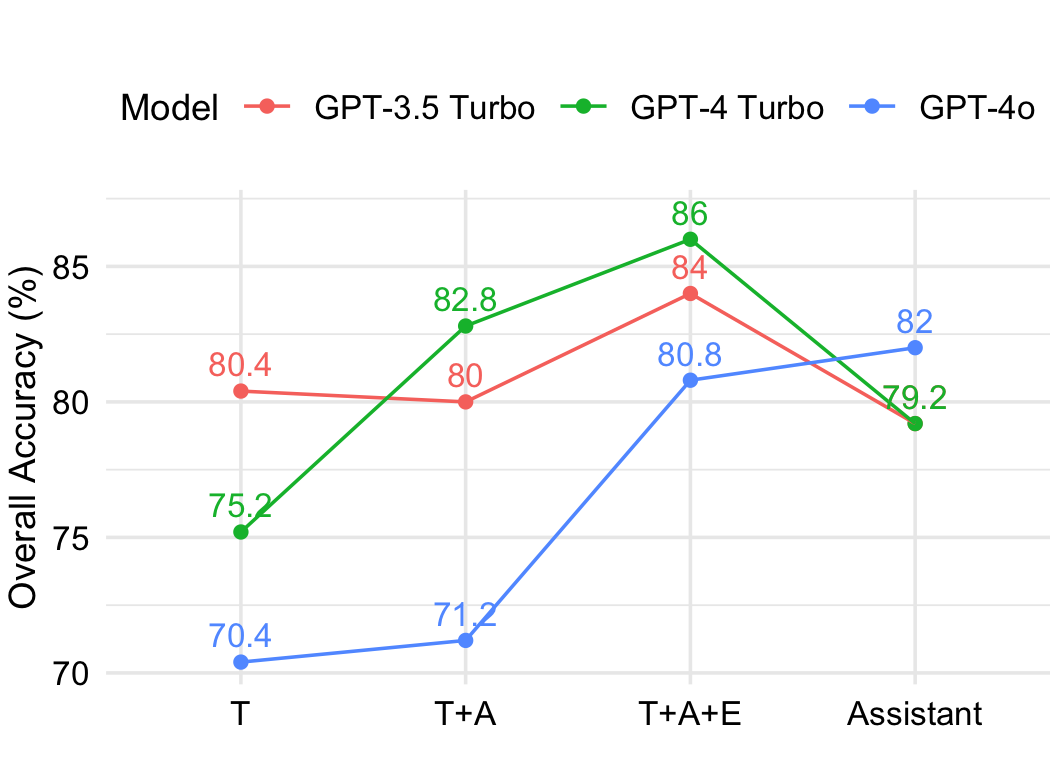}
  \caption{Two-class performance of LLMs. “T”, “A”, “E” stand for title, abstract, and excerpts, respectively.}
  \label{fig:results_2class}
\end{figure}

Considering both the rareness of Partially substantiated pairs in the dataset and their relatively low importance from a practical perspective, we then merged Partially and Fully substantiated predictions and performed a secondary analysis of model performance. Two-class performance (Figure~\ref{fig:results_2class}) clearly shows that more reference information improved LLM predictions in general.

A scrutiny of errors and explanations of LLMs revealed distinct model behaviors that are consistent with the trends in model performance. LLM explanations typically begin with a summary of the two articles followed by a comparison of them, clearly showing how a model made its prediction. When only the title of the reference is given, LLMs are forced to rely on superficial relations between the statement and the reference. GPT-3.5 Turbo tends to be permissive as long as the two articles seem to belong to the same research area. When more information is provided, GPT-3.5 Turbo would still rely on such superficial relations, and additional text from the reference that is not directly related to the statement would cause GPT-3.5 Turbo to regard the two articles as not so related. This also explains why GPT-3.5 Turbo’s performance on Unsubstantiated cases increased dramatically when more information was provided. Unlike GPT-3.5 Turbo, GPT-4 Turbo and GPT-4o tend to be much stricter in error detection, paying attention to small differences between the statement and the reference, although such differences are usually acceptable for humans since researchers frequently rephrase, abstract, and generalize statements from references, or even combine them with additional information. For instance, the statement in error example 2 of Appendix~\ref{sec:llm_error_examples} mentions copper as an example of transition metal catalysts while introducing Grignard reagents, and GPT-4 Turbo predicted this statement as Partially substantiated because the reference article about Grignard reagents does not mention copper explicitly.

Considering that hallucination can be a major threat to LLM generation, we carefully reviewed LLM-generated explanations during error analysis. No significant sign of hallucination was found.

\section{Discussion}

Scientific scholarship depends heavily on the ability to trust, verify, replicate, or refute prior research. The increasing availability of electronic publications, advanced search, and citation managers has expanded the ability of researchers to cite existing findings, yet this process is still prone to the introduction of reference errors.

This study contributes to the growing literature that seeks to utilize artificial intelligence to assist in the writing, reviewing, and publishing of scientific papers by providing a general-domain dataset of statement-reference pairs to encourage future studies in this field. We found that LLMs can detect quotation errors in scientific papers with limited context and without fine-tuning. We also quantified the relative contributions of model versions and increasing levels of context which could affect cost and speed in a production environment. During data collection, we also observed that a higher frequency of major reference errors in a paper appears to reflect citation manipulation behaviors. Given the capability of LLMs to detect quotation errors, this work may also contribute to reducing academic misconduct and fake papers that are polluting the scientific literature \citep{Sanderson24}.

Our results reveal misalignment in the comprehension of reference errors between humans and LLMs, which points the way to several promising avenues for further research. Interrogation into the features and mechanisms of quotation error detection may yield insights into which models, training corpora, or fine-tuning schemes would further improve performance. Prompt engineering (e.g. few-shot learning, multi-step chain-of-thought reasoning), ensemble methods, and other approaches are likely to demonstrate further gains in accuracy. Certain domains or types of complex statements appear more resistant to accurate classification; identifying these would help improve performance, characterize system limitations, and develop products to aid comprehensive and efficient systems for editorial and fact-checking tasks.

\section{Limitation}

Our study was limited in size and scope. During the progress of the study, new variants of Claude, Gemini, Llama, and GPT became available, some of which support a long enough context window to accept an entire reference article as input. Other models such as AI21 have more extensive token dictionaries. It is likely that the out-of-the-box performance on this task will continue to advance. 

Other limitations included our reliance on publicized and crowd-sourced datasets, the use of a simple sentence pair annotation scheme, and the treatment of all reference pairs as being equivalent despite potential multiple rationales for citation. In addition, all statement-references pairs in our dataset came from journal articles and were predominantly in natural science domains. Such characteristics limit the generalizability of our findings to papers that are published through other channels (e.g., conferences and preprint platforms) or in research domains that are underrepresented in this study. Further study would benefit from an expanded dataset labeled by multiple domain experts, and a randomly sampled and multilingual dataset indexed by year.

\section*{Acknowledgments}

The authors received no funding for this study.

\bibliography{custom}

\appendix

\section{Examples of Quotation Errors}
\label{sec:quotation_error_example}

\textbf{Example 1: Partially substantiated}\\
\textit{Reason}: Incorrect number\\
\textit{Statement}:\\
The relative topography of the origins of the renal aa. from the aorta was found to be variable, thus confirming earlier observations from angiographic studies in which the right a. arose more proximally in \underline{65\%} of cases [citation mark].\\
\textit{Reference}:\\
The ostium of the right renal a. was more cranial than the ostium of the left renal a. (\underline{53.3\%}) ...\\
\textit{Label source}: \citep{Smith20}\\
\\
\textbf{Example 2: Partially substantiated}\\
\textit{Reason}: Missing condition\\
\textit{Statement}:\\
At very high crystallinities (very low melt permeabilities), gas-generated overpressures can fracture and brecciate the solidifying mush [citation mark].\\
\textit{Reference}:\\
... for overpressure to develop, the combined crystallization expansion of plagioclase and exsolution of water is required to overcome the crystallization contraction of the other crystallizing phases... The requirement that \underline{plagioclase joins the liquidus} for crystallization overpressure to develop is significant...\\
\textit{Label source}: \citep{Smith20}\\
\\
\textbf{Example 3: Unsubstantiated}\\
\textit{Reason}: Irrelevant reference\\
\textit{Statement}:\\
There is an extremely significant positive correlation between students’ behavioral participation, cognitive participation, emotional participation, and their variables and translation performance in \underline{high school translation classrooms} [citation mark].\\
\textit{Reference}:\\
(Title) Strong negative correlation between \underline{codon usage bias and protein structural disorder} impedes protein expression after codon optimization\\
\textit{Label source}: PubPeer\\
\\
\textbf{Example 4: Unsubstantiated}\\
\textit{Reason}: Contradiction\\
\textit{Statement}:\\
We know that oxytocin is released when listening to music, and importantly, oxytocin is \underline{increased} when participating in several forms of group singing, including improvisation [citation mark].\\
\textit{Reference}:\\
We quantified mood and salivary OXT and cortisol (CORT) concentrations... Happiness was increased, and worry and sadness as well as salivary CORT concentrations were \underline{reduced}, after both choir and solo singing.\\
\textit{Label source}: Journal correction

\section{Data Collection Details}
\label{sec:data_collection_details}

Statement-reference pairs in the dataset were collected through the following channels: (1) 163 (65.2\%) pairs are from previous citation verification studies that either provided traceable examples of quotation errors or shared annotated datasets (\citealp{Lee99, Fenton00, Gosling04, Lukic04, Buchan05, Handoll15, Smith20}). Since the annotated datasets may not provide sufficient details about the specific sentence to be verified, we reviewed each citing article to locate the statement to which the label was assigned. (2) 80 (32.0\%) pairs are from PubPeer\footnote{\url{https://pubpeer.com/}}, a platform for researchers to leave comments on others' publications. While collecting annotations from this source, we cross-referenced papers retracted in 2022 and 2023 due to "concerns or issues about referencing or attributions" in the Retraction Watch Database\footnote{\url{http://retractiondatabase.org/}} with those receiving comments mentioning problematic citations on PubPeer; (3) 7 (2.8\%) pairs are from corrections, errata, and corrigenda available in the PubMed database.

Three additional inclusion criteria were applied to the dataset to ensure data quality and experiment reproducibility. First, both the citing and the reference article should have digital versions that are findable through search engines. Second, the reference should be a journal article such that text extraction from the PDF and embedding-based retrieval of text chunks are feasible. Third, the statement to be verified can be uniquely identified in the citing article. The last criterion is necessary because a reference can be cited multiple times for different statements in a paper.

\section{Prompt Template}
\label{sec:prompt_template}

\textbf{User:}\\
You are an experienced scientific writer and editor. You will be given a statement from an article that cites a reference article and information from the reference article. You will determine and explain if the reference article supports the statement.\\
\\
Specifically, choose a label from "Fully substantiated", "Partially substantiated", and "Unsubstantiated". Further explanations of the labels are as follows:\\
"Fully substantiated": The reference article fully substantiates the relevant part of the statement from the present article.\\
"Partially substantiated": According to the reference article, there is a minor error in the statement but the error does not invalidate the purpose of the statement.\\
"Unsubstantiate": The reference part does not substantiate any part of the statement. This could be because the statement is contradictory to, unrelated to, or simply missing from the reference article.\\
\\
Format your answer in JSON with two elements--"label" and "explanation". Your explanation should be short and concise.\\
\# The citing article\\
Title: \{title\}\\
Statement: \{statement\}\\
\\
\# The reference article\\
Title: \{reference\_title\}\\
Abstract: \{reference\_abstract\}\\
Excerpts: \{reference\_excerpts\}

\section{Examples of Incorrect Predictions}
\label{sec:llm_error_examples}
Abstract and excerpts of the reference were provided in the experiment but omitted here.\\
\\
\textbf{Example 1 from GPT-4 Turbo}\\
\textit{Title of the citing article}:\\
Snowball Earth climate dynamics and Cryogenian geology-geobiology\\
\textit{Statement}:\\
In East Greenland, East Svalbard, and South China [citation mark], 0.2 to 0.3 km (uncompacted) of fine-grained terrigenous strata appear.\\
\textit{Title of the reference}:\\
Reconstruction of paleo-redox conditions and early sulfur cycling during deposition of the Cryogenian Datangpo Formation in South China\\
\textit{Gold label}: Fully substantiated\\
\textit{LLM label}: Partially substantiated\\
\textit{LLM explanation}:\\
The reference article focuses on the paleo-redox conditions, sulfur cycling, and the deposition environment of the Datangpo Formation in South China during the Cryogenian period, without mentioning the specific thickness of uncompacted fine-grained terrigenous strata in East Greenland, East Svalbard, and South China. Therefore, it does not substantiate the statement regarding the thickness of terrigenous strata in these locations.\\
\\
\textbf{Example 2 from GPT-4 Turbo}\\
\textit{Title of the citing article}:\\
Transition metal-catalyzed alkyl-alkyl bond formation: Another dimension in cross-coupling chemistry\\
\textit{Statement}:\\
In early studies, primary alkyl electrophiles were coupled with alkylmagnesium reagents (Grignard reagents) in the presence of transition metal catalysts such as copper [citation mark].\\
\textit{Title of the reference}:\\
New developments in palladium catalysed cross coupling: The coupling of alkyl iodides with alkyl Grignard reagents\\
\textit{Gold label}: Fully substantiated\\
\textit{LLM label}: Partially substantiated\\
\textit{LLM explanation}:\\
The reference article supports the use of alkylmagnesium reagents (Grignard reagents) in cross-coupling reactions with alkyl electrophiles, specifically alkyl iodides, in the presence of a transition metal catalyst (palladium). However, it does not mention copper as a catalyst, focusing instead on a palladium-based system. Thus, while the reference article substantiates the general process of alkyl-alkyl bond formation using transition metal catalysts and Grignard reagents, it does not support the specific mention of copper as a catalyst.\\
\\
\textbf{Example 3 from GPT-4 Turbo}\\
\textit{Title of the citing article}:\\
Solar radiation impact on ferrofluid convection with applying electric field\\
\textit{Statement}:\\
The nanoliquids are a mixture of typical fluids with nano-powders illustrating great range of heat transfer [citation mark].\\
\textit{Title of the reference}:\\
Comprehensive evaluation of flat plate and parabolic dish solar collectors’ performance using different operating fluids and MWCNT nanofluid in different climatic conditions\\
\textit{Gold label}: Unsubstantiated\\
\textit{LLM label}: Fully substantiated\\
\textit{LLM explanation}:\\
The reference article directly supports the statement by demonstrating that the addition of nanoparticles (e.g., MWCNT) to base fluids increases the thermal conductivity of the fluid, thereby illustrating a great range of heat transfer enhancement. This is evidenced by the specific mention of increased thermal conductivity and better performance of solar systems using nanofluids.

\end{document}